\documentclass[lettersize,journal]{IEEEtran}
\usepackage{amsmath,amsfonts}
\usepackage{algorithm}
\usepackage{array}
\usepackage[caption=false,font=normalsize,labelfont=sf,textfont=sf]{subfig}
\usepackage{textcomp}
\usepackage{stfloats}
\usepackage{url}
\usepackage{verbatim}
\usepackage{graphicx}
\usepackage{cite}
\usepackage{float}
\usepackage{subfig}
\usepackage{multirow}
\usepackage{color}
\usepackage{threeparttable}
\usepackage{makecell}
\usepackage{colortbl}

\usepackage{booktabs}
\usepackage{algpseudocode}
\usepackage{amsmath}
\usepackage{amssymb}
\usepackage{subcaption}
\usepackage{multicol}
\usepackage{newfloat}
\usepackage{listings}
\floatstyle{ruled}
\newfloat{listing}{tb}{lst}{}
\floatname{listing}{Listing}
\usepackage{bibentry}

\def\onedot{. }
 
\def\ie{\emph{i.e}\onedot}

\begin{document}

\title{CrossDiff: Exploring Self-Supervised Representation of Pansharpening via Cross-Predictive Diffusion Model}

\author{Yinghui Xing,~\IEEEmembership{Member,~IEEE}, Litao Qu,  Shizhou Zhang, Kai Zhang, Yanning Zhang,~\IEEEmembership{Senior~Member,~IEEE}

\thanks{Yinghui Xing, Litao Qu, Shizhou Zhang, Yanning Zhang are with the Shaanxi Provincial Key Laboratory of Speech and Image Information Processing, and the National Engineering Laboratory for Integrated Aerospace-GroundOcean Big Data Application Technology, School of Computer Science, Northwestern Polytechnical University, Xi’an 710072, China. Yinghui Xing is also with the Research
Development Institute of Northwestern Polytechnical University in
Shenzhen, Shenzhen 518057, China. (e-mail: xyh\_7491@nwpu.edu.cn).
Kai Zhang is with the School of Information Science and Engineering, Shandong Normal University, 250358, Jinan
}
}

\maketitle

\begin{abstract}
Fusion of a panchromatic (PAN) image and corresponding multispectral (MS) image is also known as pansharpening, which aims to combine abundant spatial details of PAN and spectral information of MS.
Due to the absence of high-resolution MS images, available deep-learning-based methods usually follow the paradigm of training at reduced resolution and testing at both reduced and full resolution. When taking original MS and PAN images as inputs, they always obtain sub-optimal results due to the scale variation. In this paper, we propose to explore the self-supervised representation of pansharpening by designing a cross-predictive diffusion model, named CrossDiff. It has two-stage training. In the first stage, we introduce a cross-predictive pretext task to pre-train the UNet structure based on conditional DDPM, while in the second stage, the encoders of the UNets are frozen to directly extract spatial and spectral features from PAN and MS, and only the fusion head is trained to adapt for pansharpening task. Extensive experiments show the effectiveness and superiority of the proposed model compared with state-of-the-art supervised and unsupervised methods. Besides, the cross-sensor experiments also verify the generalization ability of proposed self-supervised representation learners for other satellite's datasets. We will release our code for reproducibility.

\end{abstract}

\begin{IEEEkeywords}
Heterogeneous images, change detection, auto-encoder, pseudo-label, progressive.
\end{IEEEkeywords}

\section{Introduction}
\IEEEPARstart{P}{ansharpening} refers to fusing a low-spatial-resolution multispectral (LRMS) image with a single-band high-spatial-resolution panchromatic (PAN) image to obtain a high spatial and spectral resolution MS (HRMS) image.
Owing to the technological and physical limitations of imaging devices, remote sensing images obtained by satellite sensors always have a trade-off between spatial and spectral resolution~\cite{wang2016adaptive}.
Therefore, most of available satellites carry two types of sensors, \ie, an MS sensor and a PAN sensor, to acquire LRMS and PAN images simultaneously. Researchers can then combine them through pansharpening technology.

\begin{figure}[h]
  \centering
  \includegraphics[width=\linewidth]{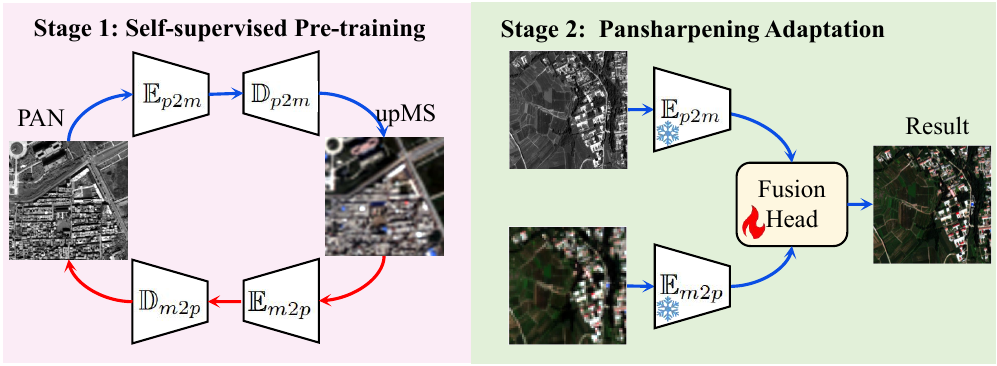}
  \caption{Training of our method. Stage 1: Pre-train the UNet structure with a cross-predictive pretext task based on conditional DDPM.
  Stage 2: Train the fusion head to adapt for pansharpening task.}
  \label{fig:overview}
\end{figure}

Benefiting from the availability of large-scale remote sensing imagery, numerous deep-learning (DL)-based methods achieve great success in pansharpening. According to the base models they used, these methods can be roughly divided into convolutional neural networks (CNNs) based~\cite{machinelearing_pansharpening,yang2017pannet,masi2016PNN}, generative adversarial networks (GANs) based~\cite{liu2020psgan,xu2023upangan,ma2020panGan}, and Transformer based~\cite{zhou2022panformer,bandara2022hypertransformer,zhou2022AAAI_invertable_transformer}. CNNs-based methods mainly extract spatial and spectral features layer-by-layer. They take pansharpening as a regression task, and are usually supervised by a reconstruction loss. GANs-based models utilize generators to fuse MS and PAN images and the discriminators to adversarially train the model for the generation of high-fidelity fused products. To model the long-range dependency of images, researchers introduce the Transformer architecture to pansharpening~\cite{zhou2022panformer}. These models can learn either the self-similarity in a single image~\cite{bandara2022hypertransformer} or the interactive information between two kinds of modality images~\cite{zhou2022AAAI_invertable_transformer}.

Most of available DL-based pansharpening methods are within the supervised-learning framework. However, there are no high-resolution MS images that can be taken as groundtruths. Therefore, researchers train the models at reduced resolution, where the training samples are prepared according to Wald's protocol~\cite{wald1997wald}, \ie, downsample PAN and MS to take them as inputs, and the original MS images are treated as references, which are actually pseudo-groundtruths~\cite{zhang2023p2sharpen}. Nevertheless, due to the scale variation~\cite{ciotola2022pansharpening,ma2020panGan}, models trained at reduced resolution are unsuitable for pansharpening at the original images, resulting in inferior performance at full resolution~\cite{xu2023upangan}.

Recently, researchers devote to developing unsupervised pansharpening methods.
Most of them~\cite{luo2020unsupervisedCNN,ciotola2022ZPNN} concentrate only on the design of unsupervised loss function, however their effectiveness also depends on the
features extracted~\cite{van2017neural}.
We argue that the scale dependency issue can be alleviated from the perspective of not only  unsupervised loss function, but also a good representation learner.
Recent works utilize generative models to act as representation learners~\cite{van2017neural,donahue2019large,dhariwal2021DDPMbeatgans}.
Among them, Denoising Diffusion Probabilistic Model (DDPM)~\cite{ho2020DDPM}  outperforms other alternative generative models and provides
more semantically-valuable pixel-wise representations~\cite{baranchuk2021pre_train_DDPM}, which highlights the potential of using the state-of-the-art DDPM as strong unsupervised representation learners. Based on this insight, we build model upon DDPM to explore its representation ability to pansharpening.

Furthermore, available DL-based methods have limited generalization ability, where the model trained on a specific dataset always performs unsatisfactorily on datasets collected from other satellite sensors. This further arises a question: how can we develop a unified representation learner based on DDPM to make it accountable for the extraction of general spatial and spectral features, regardless of the distinctiveness of sensors' attributes.

For pansharpening task, PAN and MS images have distinctive spatial and spectral characteristics respectively, which naturally constitutes a self-supervised pretext task that one can predict high-resolution PAN image from low-resolution MS image, and inversely predict spectral-appealing MS image from spectral-scarce PAN image.
In this paper, we embed pansharpening into diffusion model to construct a new self-supervised paradigm, dubbed CrossDiff.
As Figure~\ref{fig:overview} shows, the model consists of two stage training. In the first stage, the self-supervised pre-training is conducted based on DDPM to obtain a noise predictor with UNet structure~\cite{ronneberger2015unet}. Afterwards, the frozen encoders act as spatial-spectral feature representation learners to fuse MS and PAN images with a tunable fusion head.
By introducing a cross-predictive diffusion process,
CrossDiff outperforms state-of-the-art unsupervised pansharpening methods and has strong generalization ability. 

The main contributions are summarized as follows:

\begin{itemize}
    \item We design a new two-stage pansharpening paradigm to explore the potential of DDPM to the self-supervised spatial and spectral features extraction.
    \item A cross-predictive diffusion process is introduced to pre-train the spatial and spectral representation learners. Following the process of DDPM, the effective training objectives encourage the pretext task to explicitly learn spatial and spectral diffusion latent.
    \item By freezing the pre-trained models and only tuning the fusion head, our proposed CrossDiff performs well at both full and reduced resolution, and has strong cross-sensor generalization ability.
\end{itemize}

\begin{figure*}[h]
  \centering
  \includegraphics[width=\linewidth]{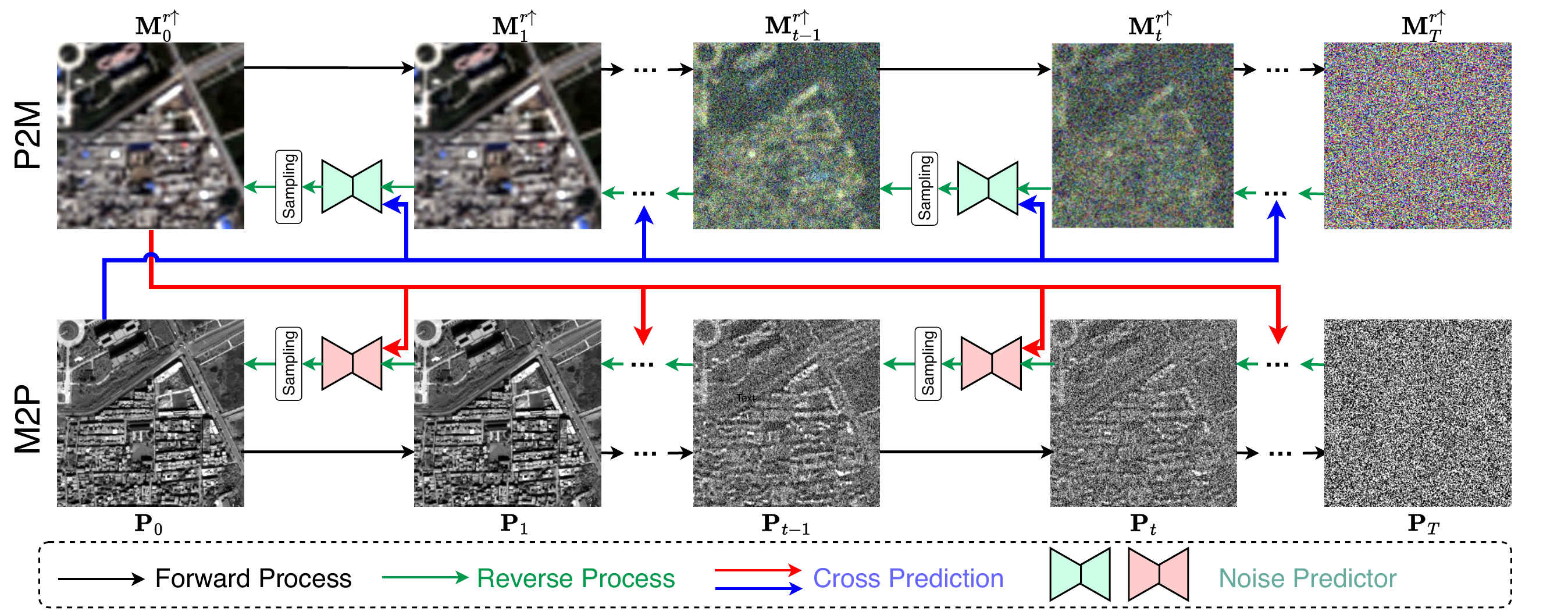}
  \caption{Framework of proposed cross-predictive pretext task, where PAN and upsampled MS images are cross predicted from each other through the reverse process.}
  \label{fig:pretext_task}
\end{figure*}

\section{Related Works} \label{sc:related}

\subsection{Pansharpening with Deep Generative Model.}
The development of deep generative models, especially GANs, boosts their applications to pansharpening.
PSGAN~\cite{liu2020psgan} was a first attempt to produce high-quality results under the framework of GANs.
Taking advantages of multiple discriminators in GANs, ~\cite{gastineau2021generative} designed a bi-discriminator GANs method for pansharpening, where one of the discriminator was optimized for texture preservation and the other was for color preservation. Though effective, they are trained at reduced resolution, which are not suitable for images at original scale due to the scale-variation.
Therefore, \cite{ma2020panGan} trained a GANs-based network directly on the full-resolution images using two discriminators to respectively constrain the spatial and spectral consistency of fusion results. UCGAN~\cite{zhou2022UCGAN} extracted spatial and spectral information based on a two-stream generator, and introduced a cycle-consistency loss to further improve the pansharpening performance. Different from these unsupervised methods, ~\cite{diao2022zergan} designed a ``training is fusion'' framework based on multiscale GANs, which got rid of the dependency of large amounts of training data.
However, GANs are easily caught in unstable training. As another type of generative models, diffusion model has attracted much attention in AIGC, but it is rarely studied in pansharpening. We propose a diffusion model based pansharpening method to generate images with more spatial details than those GANs-based ones.

\subsection{Self-Supervised Representation Learning.}
Self-supervised learning (SSL) involves learning powerful representations by replacing human annotation with pretext tasks derived from data alone~\cite{he2020MOCO, ziegler2022self}. Colorization~\cite{larsson2017colorization_as_proxy,zhang2017split_brain} and image super-resolution~\cite{wang2023energy} are two commonly used pretext tasks in SSL, which enable networks to learn structure, context, and semantic features from images by generating realistic colorful or high-resolution images. These learned features can then be transferred to downstream tasks using transfer learning.
\cite{wang2022multifocus} introduced image super-resolution (SR) into multi-focus image fusion, where the model trained on SR can be directly used without any fine-tuning.
As for the pansharpening, \cite{ozcelik2020rethinking} presented a colorization-based SSL framework for pansharpening, which took grayscale transformed MS image as input, and trained the model to learn the colorization of it.


\section{Preliminaries} \label{sc:preliminaries}
\subsection{Denoising Diffusion Probabilistic Model (DDPM).}
In this section, we provide a brief review of DDPM. One can refer to~\cite{ho2020DDPM,nichol2021improvedDDPM} for more details.

DDPM is a class of generative model that learns a data distribution by denoising the noisy images.
DDPM has achieved excellent results in image super-resolution~\cite{saharia2022SR3}, text-to-image generation~\cite{saharia2022imagen,ramesh2022dalle2,rombach2022stable_diffusion}, semantic segmentation~\cite{baranchuk2021pre_train_DDPM}, and change detection~\cite{bandara2022ddpm-cd}. It defines a Markov chain of diffusion steps to progressively add random noises to data (forward process) and then learns to reverse the diffusion process to reconstruct desired data samples from the pure noise (reverse process).

Given a sequence of positive noise scales $0<\alpha_{1},\alpha_{2},\cdots,\alpha_{T}<1$, DDPM defines a discrete Markov chain $\{x_{1},x_{2},\dots,x_{T}\}$ for each training data point $x_{0}\sim q(x_0)$, such that $q(x_{t}\mid x_{t-1})=\mathcal{N}(x_{t};\sqrt{\alpha_{t}}x_{t-1},(1-\alpha_{t})\mathbf{I})$. Therefore, we obtain     $q(x_{t}\mid x_{0})=\mathcal{N}(x_{t};\sqrt{\gamma_{t}}x_{0},(1-\gamma_{t})\mathbf{I})$ by defining $\gamma_{t} =  {\textstyle \prod_{i=1}^{t}}\alpha_{i}$. As time $\mathinner{t}$ increases, $x_{t}$ is gradually corrupted by Gaussian noise, and when the total number of diffusion step $\mathinner{T}$ is large enough, $x_{T}$ will be equivalent to an isotropic Gaussian distribution $\mathcal{N}(\mathbf{0},\mathbf{I})$.

The reverse process of DDPM is to recover the real data distribution $q(x_0)$ by progressively denoising from the Gaussian noise. Using the Bayes theorem, we know that the posterior $q(x_{t-1}\mid x_{t})$ also follows a Gaussian distribution, whose mean and variance can be calculated through the combination of $x_t$ and $x_0$, together with $\alpha_t$ and $\gamma_t$. However, the real data distribution is unknown, and we cannot exactly obtain the mean value of the posterior. Therefore, we need to approximate it using $p_{\theta}(x_{t-1}\mid x_{t}) = \mathcal{N}(x_{t-1};\mu_{\theta}(x_{t},t),\sigma_{\theta}^2\mathbf{I})$. To obtain the mean $\mu_{\theta}$ and the variance $\sigma_{\theta}^2$, DDPM is trained with a re-weighted variant of the evidence lower bound. A more straightforward objective in the training phase is to obtain the noise predictor $\epsilon _{\theta }(x_{t},t)$ by minimizing $E_{t\sim [1,T],x_{0}\sim q(x_{0}),\epsilon\sim \mathcal{N}(0,1)}\left \| \epsilon _{\theta }(x_{t},t)-\epsilon  \right\| _{2}$~\cite{ho2020DDPM}. Starting from $x_T \sim \mathcal{N}(\mathbf{0},\mathbf{I})$, the sampled $x_{t-1}$ can be calculated by: $x_{t-1}=\frac{1}{\sqrt{\alpha _{t}} }(x_{t}-\frac{1-\alpha _{t}}{\sqrt{1-\gamma _{t}} }\epsilon _{\theta }(x_{t},t) ) + \sigma_{\theta}z_t$, where $z_t \sim \mathcal{N}(\mathbf{0},\mathbf{I})$. The variance $\sigma_{\theta}^2$ can be directly set to $1-\alpha _{t}$ or $\frac{(1-\alpha_{t})(1-\gamma_{t-1})}{1-\gamma_{t}}$.

\subsection{Conditional DDPM.}
The conditional DDPM feeds conditional information to establish the conditional posterior distribution $p_{\theta}(x_{t-1}\mid x_{t},y)$, where $y$ denotes the condition. The conditions can be class labels~\cite{nichol2021improvedDDPM}, images~\cite{saharia2022SR3} or text embeddings~\cite{saharia2022imagen}, and can also be fed to the noise predictor in a flexible way. The formulation of conditional DDPM is similar to DDPM, and we only need to modify $\epsilon_{\theta}(x_{t},t)$ to $\epsilon_{\theta}(x_{t},y,t)$ in the optimization and sampling process.

\section{Methodology} \label{sc:method}

For the sake of simplicity, we first provide some notations used in this paper.
We use $\mathbf{P}\in \mathbb{R}^{W\times H \times 1}$ and $\mathbf{M} \in \mathbb{R}^{\frac{W}{r}\times \frac{H}{r} \times C}$ to denote the PAN and MS images, where $W$ and $H$ are the width and height of PAN. $C$ is the band number of MS, and $r$ is the spatial resolution ratio of MS to PAN. The fusion result can then be represented by $\mathbf{FMS}\in \mathbb{R}^{W\times H\times C}$. Note that in our paper, $r\downarrow$ and $r\uparrow$ denote the operations that down-sample or up-sample images $r$ times, respectively.

\subsection{Overview}
The whole training process of proposed method is
demonstrated in Figure~\ref{fig:overview}.
CrossDiff has a self-supervised pre-training stage and a pansharpening adaptation stage.
In the self-supervised pre-training stage, a cross-predictive diffusion model is designed, which consists of a P2M branch and a M2P branch, aiming to constrain the extraction of spatial and spectral features. 
As Figure~\ref{fig:pretext_task} shows, P2M takes PAN image $\mathbf{P}_0$ as condition to guide the model to reconstruct MS image through optimization and sampling. Similarly, M2P takes upsampled MS image $\mathbf{M}^{r\uparrow}_0$ as guidance to produce PAN.
The training objective of this stage builds upon the modeled noises.
After the cross-predictive self-supervised pre-training, we obtain two noise predictors with UNet structure. Their encoders, \ie $\mathbb{E}_{p2m}$ and $\mathbb{E}_{m2p}$, are frozen and then taken as feature extractors to extract spatial and spectral features in the pansharpening adaptation stage. In the second stage, we only need to train the fusion head to make it adapt for pansharpening task. During the inference phase, we use the pre-trained encoders and the tuned fusion  head to obtain the final fusion results.

\begin{figure}[tbp]
  \centering
  \includegraphics[width=\linewidth]{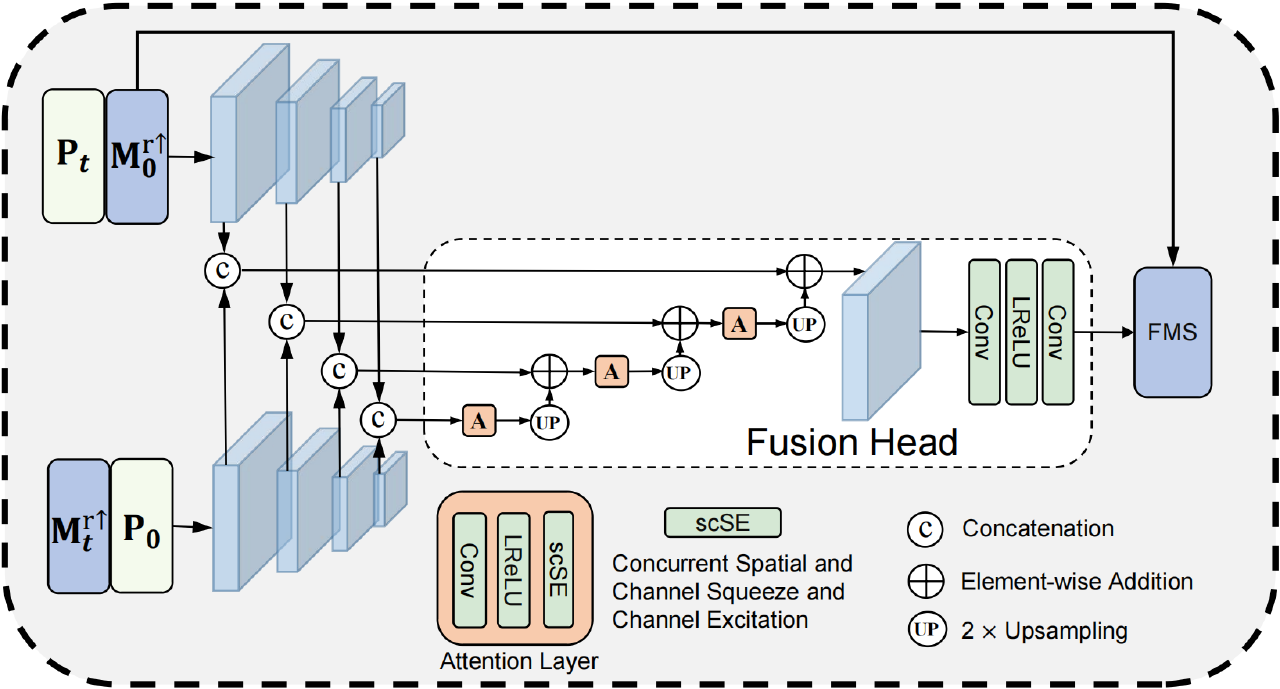}
  \caption{The pansharpening adaptation stage, where the spatial and spectral features are extracted from the pre-trained encoders, and concatenated to be taken as inputs to the fusion head.
  The fusion result FMS is then obtained. }
  \label{fig:fusion_module}
\end{figure}

\subsection{Self-Supervised Pre-Training}\label{sc:pretext_task}
The self-supervised pre-training is realized through a cross-predictive diffusion process shown in Figure~\ref{fig:pretext_task}. It consists of a P2M process and an M2P diffusion process. P2M and M2P are modeled by two conditional DDPMs, where PAN and MS images are treated as conditions respectively to reconstruct each other. In the following, we introduce the cross-predictive diffusion model from the view of forward and reverse processes.

\textbf{Forward Diffusion Process.}
Starting from PAN image $\mathbf{P}_0$ or upsampled MS image $\mathbf{M}^{r\uparrow}_0$, forward diffusion process progressively adds noises to obtain a series of images $\{\mathbf{P}_1, \mathbf{P}_2,\cdots,\mathbf{P}_T \}$ or $\{\mathbf{M}^{r\uparrow}_1, \mathbf{M}^{r\uparrow}_2,\cdots,\mathbf{M}^{r\uparrow}_T \}$. This process is formulated by
\begin{equation}
\begin{aligned}
  q(\mathbf{P}_t\mid \mathbf{P}_{t-1}) &=\mathcal{N}(\mathbf{P}_{t};\sqrt{\alpha_{t}}\mathbf{P}_{t-1},(1-\alpha_{t})\mathbf{I}),
   \\ q(\mathbf{M}^{r\uparrow}_t\mid \mathbf{M}^{r\uparrow}_{t-1})
 &=\mathcal{N}(\mathbf{M}^{r\uparrow}_{t};\sqrt{\alpha_{t}}\mathbf{M}^{r\uparrow}_{t-1},(1-\alpha_{t})\mathbf{I}),
\end{aligned}\label{eq:1}
\end{equation}
where $\alpha_t~ (t=1,2,\cdots,T)$ represents the noise scale, and we use cosine noise scheduler~\cite{nichol2021improvedDDPM} to assign its value. Note that when $T$ is large enough, $\mathbf{P}_T$ and $\mathbf{M}^{r\uparrow}_T$ can be isotropic Gaussian noise $\epsilon$.

\textbf{Cross-Predictive Reverse Process.} The reverse process is a cross-predictive task, where both P2M and M2P start from the Gaussian noise, and take UNets as noise predictors to sample and reconstruct image at each time step. Like conditional DDPM~\cite{saharia2022SR3}, PAN and MS are taken as inputs of UNets
for P2M and M2P respectively, to reconstruct each other.

Specifically, P2M takes PAN image $\mathbf{P}_0$ as condition to produce MS image.
Considering the $t$-th time step, the noise predictor $\epsilon_{\varphi}(\cdot)$ of P2M branch takes $\mathbf{P}_0$, $\mathbf{M}^{r\uparrow}_{t}$ and the time embedding $t$ as inputs to approximate the noise
\begin{equation}  \label{eq:2}
min \left \| \epsilon_{\varphi}(\mathbf{M}_{t}^{r\uparrow},\mathbf{P}_0,t) -\epsilon \right \|_{2}^{2},
\end{equation}
where $\epsilon_\varphi(\mathbf{M}^{r\uparrow}_t,\mathbf{P}_0,t)$ is the noise predicted by UNet in $t$-th time step of P2M, and $\epsilon$ is an isotropic Gaussian noise.

After obtaining $\epsilon_\varphi(\mathbf{M}^{r\uparrow}_t,\mathbf{P}_0,t)$, we derive its mean $\mu_{\varphi}(\mathbf{M}_{t}^{r\uparrow},\mathbf{P}_{0},t)$ and the variance $ \sigma_{\varphi}^{2}$ using

\begin{align}\label{eq:3}
    \mu_{\varphi}(\mathbf{M}_{t}^{r\uparrow},\mathbf{P}_0,t)&=\frac{1}{\sqrt{\alpha _{t}} }(\mathbf{M}_{t}^{r\uparrow}-\frac{1-\alpha _{t}}{\sqrt{1-\gamma _{t}} }\epsilon _{\varphi }(\mathbf{M}_{t}^{r\uparrow},\mathbf{P}_0,t) ), \nonumber
   \\
   \sigma_{\varphi}^{2}&=\frac{(1-\alpha_{t})(1-\gamma_{t-1})}{1-\gamma_{t}}.
\end{align}

 Then, we can sample $\mathbf{M}_{t-1}^{r\uparrow}$ according to the mean and variance. Under the $T$-th iteration, we obtain the reconstructed MS image $\mathbf{M}_{0}^{r\uparrow}$. Note that the reconstruction of MS in this process starts from a Gaussian noise, and only takes grayscale PAN images as prior, therefore, we call it P2M process. We think P2M process enables the model to complement spectral information like the way of widely used self-supervised colorization pretext~\cite{zhang2017split_brain,larsson2017colorization_as_proxy}.

As opposed to P2M, M2P takes up-sampled MS image $\mathbf{M}_0^{r\uparrow}$ as condition to generate PAN image. The UNet structure is also taken as noise predictor to approximate the added noise $\epsilon_\theta(\mathbf{P}_t,\mathbf{M}^{r\uparrow}_0,t)$ at $t$-th time step. The optimization objective is
\begin{equation}  \label{eq:4}
min \left \| \epsilon_{\theta}(\mathbf{P}_t,\mathbf{M}^{r\uparrow}_0,t) -\epsilon \right \|_{2}^{2},
\end{equation}
and the mean and variance can be derived by
\begin{align}
     \mu_{\theta}(\mathbf{P}_t,\mathbf{M}_{0}^{r\uparrow},t)&=\frac{1}{\sqrt{\alpha _{t}} }(\mathbf{P}_{t}-\frac{1-\alpha _{t}}{\sqrt{1-\gamma _{t}} }\epsilon _{\theta }(\mathbf{P}_t,\mathbf{M}_{0}^{r\uparrow},t)), \nonumber\\
   \sigma_{\theta}^{2}&=\frac{(1-\alpha_{t})(1-\gamma_{t-1})}{1-\gamma_{t}}.
\end{align}

Similarly, we sample $\mathbf{P}_{t-1}$, and finally obtain the reconstructed PAN image $\mathbf{P}_0$. M2P takes low-spatial resolution MS images as inputs to produce high-spatial resolution PAN images, and the model used in this process can well learn the spatial details like the image super-resolution task~\cite{saharia2022SR3,wang2022multifocus}.

\textbf{Noise Predictor.}
DDPM~\cite{ho2020DDPM} introduced a UNet as noise predictor, which has been found to have improved sample quality~\cite{jolicoeuradversarial,nichol2021improvedDDPM}.
In this paper, we use the simplified UNet architecture for P2M and M2P process, which has a stack of residual layers and downsampling convolutions, followed by a stack of residual layers with upsampling operations. The features with the same spatial size are connected by skip connections. In order to train a time-dependent denoising model, diffusion time step $t$ is specified in the way of Transformer's sinusoidal position encoding to each residual block. More detailed architecture of the noise predictor can be found in the supplementary material.

\subsection{Pansharpening Adaptation }
After the self-supervised pre-training stage, we obtain the pre-trained spatial and spectral feature extractors, \ie, $\mathbb{E}_{m2p}$ and $\mathbb{E}_{p2m}$. They are frozen in the pansharpening adaptation stage to extract spatial and spectral features.

\textbf{Attention-Guided Fusion Head.}
To make the model adapt for pansharpening task, we design an attention-guided fusion head. As shown in Figure~\ref{fig:fusion_module},
the fusion head takes concatenated features of PAN and up-sampled MS at different scales as inputs.
The features at lower scale pass through the attention layer and are up-sampled and then added to next attention layer to generate the attentive features of the higher scale. The attention layer we used is similar with~\cite{bandara2022ddpm-cd}, which contains a convolutional layer with LeaklyReLU, followed by an scSE layer~\cite{roy2018scSE}.

After the attention-guided fusion, we use two convolutional layers to reconstruct the fused features.
Like PanNet~\cite{yang2017pannet}, we also learn
the residual map, and obtain the final fusion result
by combining the residuals with the up-sampled MS image $\mathbf{M}^{r\uparrow}_0$.

\textbf{Loss function.}
The self-supervised pre-training is realized based on the predicted noises, \ie, Equation~(\ref{eq:2}) and Equation~(\ref{eq:4}), in the first stage. While in the pansharpening adaptation stage, the training of fusion head is achieved through an unsupervised loss function~\cite{zhou2022UCGAN,supervised-unsupervised}.

The loss function contains a spectral term, a spatial term, and a QNR term,
\begin{equation}\label{eq:5}
    \mathcal{L}_{full} = \mathcal{L}_{QNR}+\lambda(\mathcal{L}_{spa}+\mathcal{L}_{spe}),
\end{equation}
where $\lambda$ is a balance parameter, which is set to be 0.01. The loss terms are defined as follows:
\begin{align}\label{eq:6}
     \mathcal{L}_{spa} = & \left \|getHp(\mathbf{P})-getHp(Mean(\mathbf{FMS})) \right \|_{1} \nonumber \\
    &+(1-SSIM(getHp(\mathbf{P}),getHp(Mean(\mathbf{FMS})))), \nonumber \\
    \mathcal{L}_{spe} = & \left \| \mathbf{M}^{r\uparrow}-GS(\mathbf{FMS}) \right \|_{1} \nonumber\\
     &+(1-SSIM(\mathbf{M}^{r\uparrow},GS(\mathbf{FMS}))), \nonumber \\
     \mathcal{L}_{QNR} & =  1-QNR(\mathbf{M},\mathbf{P},\mathbf{FMS}),
\end{align}
where $Mean(\cdot)$ and $getHp(\cdot)$ denote the channel average pooling operation and the high-frequency filtering operation~\cite{yang2017pannet}. $GS(\cdot)$ and $SSIM(\cdot)$ represent the Gaussian filtering operation and the Structural Similarity Index Measure (SSIM)~\cite{wang2004SSIM}. $QNR(\cdot,\cdot,\cdot)$ is calculated according to~\cite{alparone2008qnr}.

\begin{figure*}[htbp]
  \begin{tabular}{{c}{c}{c}{c}{c}{c}}
  \includegraphics[width=0.16\textwidth]{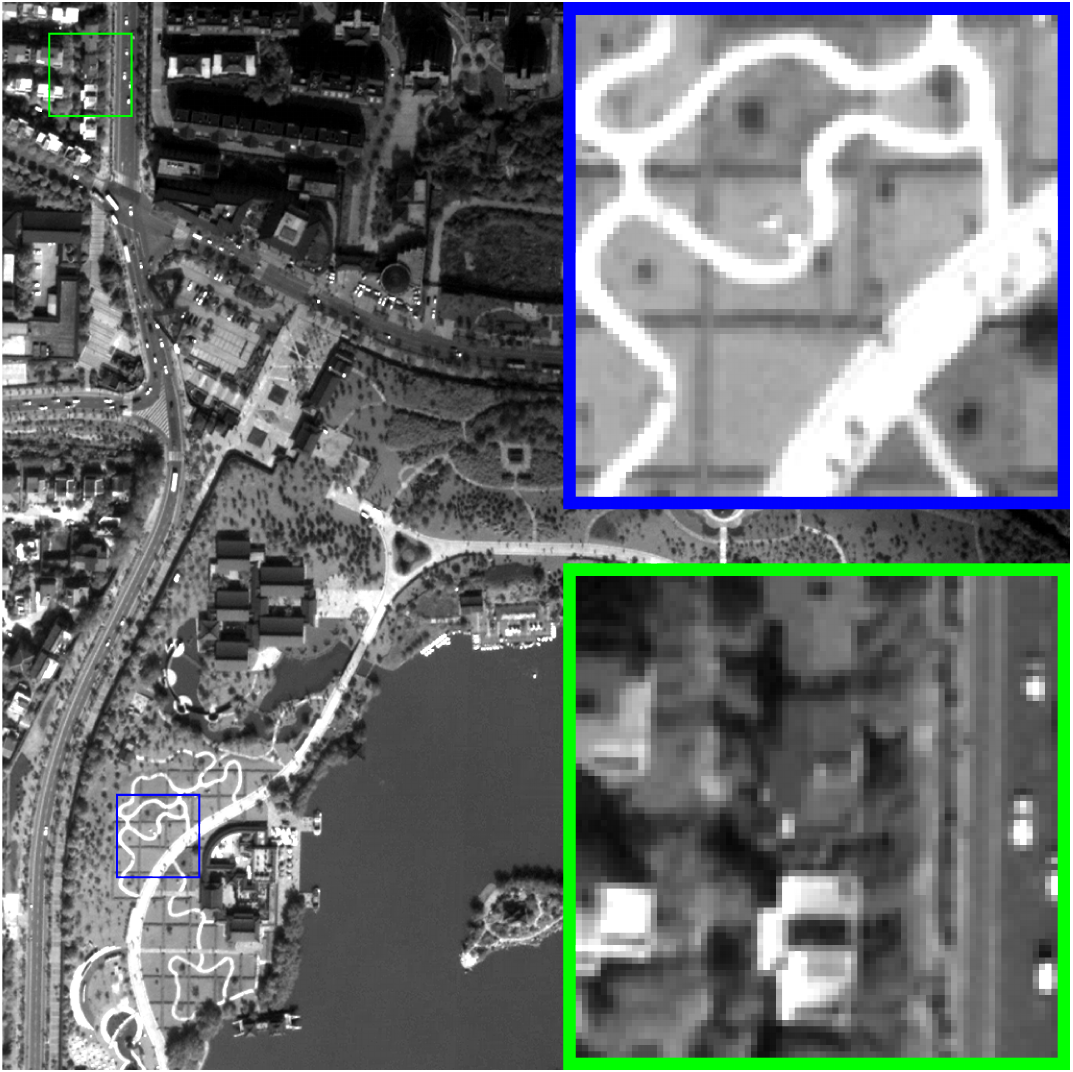} &
  \hspace{-4mm}
  \includegraphics[width=0.16\textwidth]{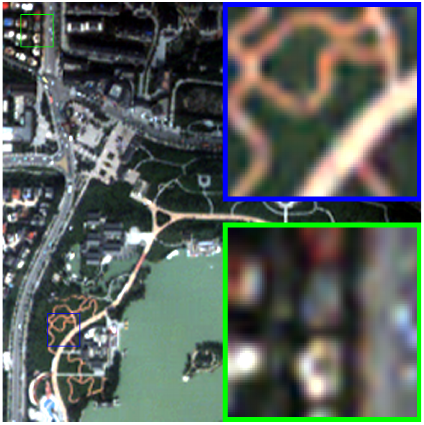} &
  \hspace{-4mm}
  \includegraphics[width=0.16\textwidth]{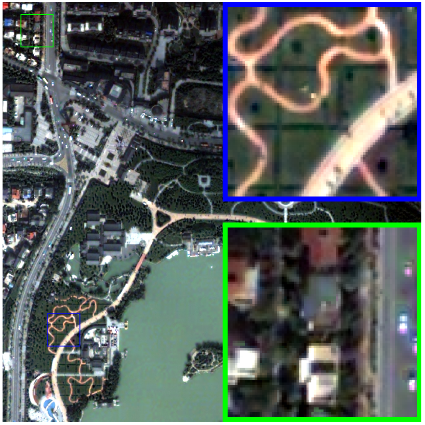} &
  \hspace{-4mm}
   \includegraphics[width=0.16\textwidth]{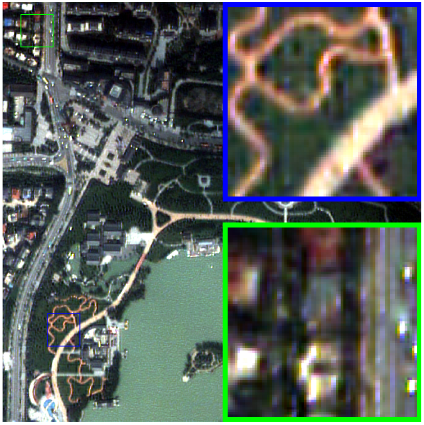}&
  \hspace{-4mm}
  \includegraphics[width=0.16\textwidth]{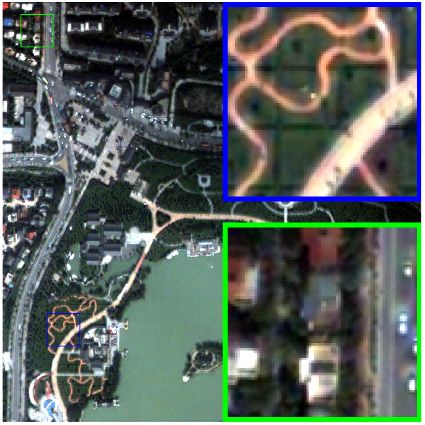}&
    \hspace{-4mm}
    \includegraphics[width=0.16\textwidth]{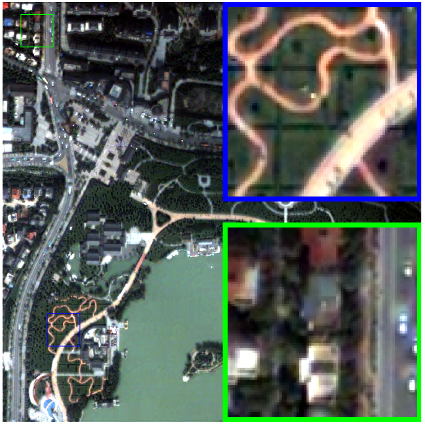}
\\
  PAN &  \hspace{-4mm}EXP &   \hspace{-4mm}BTH & \hspace{-4mm}C-GSA  &\hspace{-4mm}MTF-GLP-HPM-R&\hspace{-4mm}MTF-GLP-FS
    \\
  \includegraphics[width=0.16\textwidth]{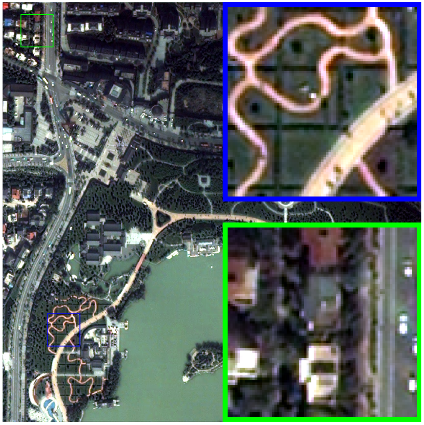}&
  \hspace{-4mm}
   \includegraphics[width=0.16\textwidth]{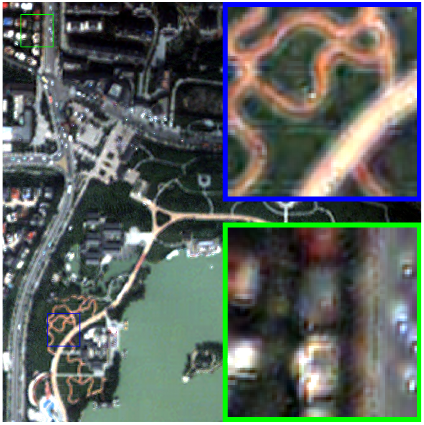}&
  \hspace{-4mm}
  \includegraphics[width=0.16\textwidth]{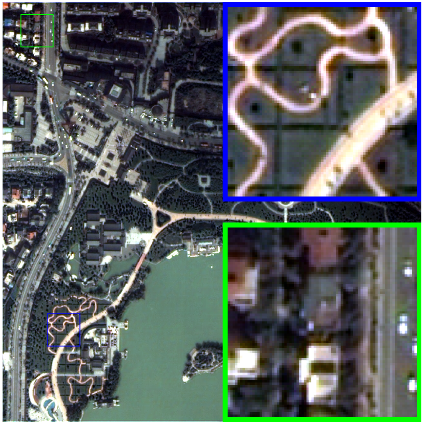}  &
   \hspace{-4mm}
  \includegraphics[width=0.16\textwidth]{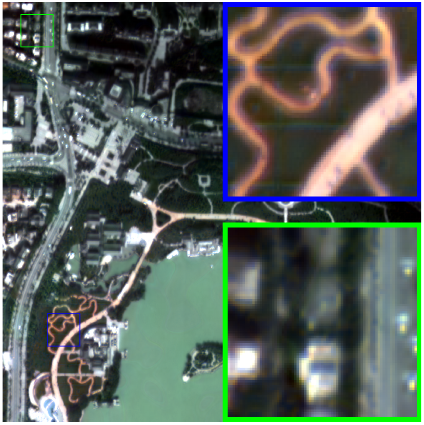}&
  \hspace{-4mm}
  \includegraphics[width=0.16\textwidth]{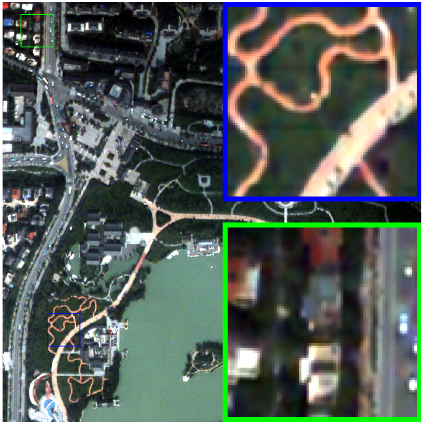}&
    \hspace{-4mm}
  \includegraphics[width=0.16\textwidth]{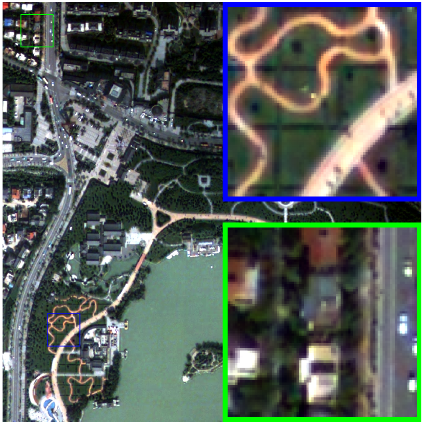}
  \\
 \hspace{-4mm}  ZPNN &\hspace{-4mm}  UCGAN&\hspace{-4mm} LDPNET& \hspace{-4mm}GDD& \hspace{-4mm}CrossDiff$^*$& \hspace{-4mm}CrossDiff \\
  \end{tabular} \normalsize
  \caption{Fusion results on QuickBird (QB) dataset at the full resolution. One can zoom in for more details.}
  \label{full_qb}
\end{figure*}

\begin{table*}
\centering
\begin{threeparttable}
	\caption{Evaluation indexes of different methods on QuickBird (QB) and WorldView-2 (WV-2) datasets at full resolution.}
		\begin{tabular}{ccccc|cccc}
			\toprule
  Dataset&\multicolumn{4}{c}{QB}&\multicolumn{4}{c}{WV-2}\\
  \midrule
           Indices&$D_{\lambda}$&$D_{s}$&HQNR&Time~(s)&$D_{\lambda}$&$D_{s}$&HQNR&Time~(s)\\
           \midrule
           Ideal value&0&0&1&0&0&0&1&0\\
            \midrule
                BT-H &0.072$\pm$0.020&0.142$\pm$0.066&0.782$\pm$0.132&0.26&0.098 $\pm$0.087 &0.154$\pm$0.083 &0.703$\pm$0.111&\underline{0.24}\\
                C-GSA&0.070$\pm$0.032&0.146$\pm$0.070&0.777$\pm$0.135&3.95&0.073$\pm$0.052&0.139$\pm$0.072&0.722$\pm$0.097&3.93\\
                MTF-GLP-HPM-R&0.089$\pm$0.028&0.128$\pm$0.032&0.843$\pm$0.042&0.45&0.076$\pm$0.056&0.099$\pm$0.085&0.857$\pm$0.104&0.79\\
                MTF-GLP-FS&0.090$\pm$0.029&0.133$\pm$0.033&0.835$\pm$0.043&0.46&0.083$\pm$0.053&0.103$\pm$0.082&0.857$\pm$0.101&0.84\\
                ZPNN&0.143$\pm$0.051&0.175$\pm$0.048&0.755$\pm$0.071&14.40&0.129$\pm$0.089&0.087$\pm$0.057&0.822$\pm$0.076&31.26\\
                UCGAN&\underline{0.022$\pm$0.020}&0.091$\pm$0.018&0.854$\pm$0.022&\underline{0.22}&\textbf{0.012$\pm$0.022}&0.062$\pm$0.031&0.859$\pm$0.047&0.27\\
                LDPNET&0.166$\pm$0.044&0.231$\pm$0.072&0.609$\pm$0.136&\textbf{0.01}&0.070$\pm$0.049&0.143$\pm$0.069&0.669$\pm$0.124&\textbf{0.01}\\
                GDD&0.042$\pm$0.025&0.042$\pm$0.014&0.824$\pm$0.082&355.79&0.050$\pm$0.044&0.059$\pm$0.026&0.795$\pm$0.165&493.60\\
                \midrule
                CrossDiff$^*$&0.029$\pm$0.018&\underline{0.020$\pm$0.010}&\underline{0.926$\pm$0.025}&0.81&0.018$\pm$0.007&\underline{0.040$\pm$0.014}&\underline{0.874$\pm$0.031}&1.65\\
                CrossDiff&\textbf{0.020$\pm$0.016}&\textbf{0.015$\pm$0.005}&\textbf{0.946$\pm$0.015}&0.81&\underline{0.014$\pm$0.008}&\textbf{0.025$\pm$0.006}&\textbf{0.899$\pm$0.045}&1.65\\
\bottomrule
	\end{tabular}
 \label{tb:full}
 \small
 $^a$ CrossDiff$^*$ denotes the cross-sensor pre-trained model, \ie, pre-trained on other satellite datasets.
 \end{threeparttable}
\end{table*}

\begin{table*}
\begin{threeparttable}
	\centering
	\caption{Evaluation indexes of different methods on QuickBird (QB) and WorldView-2 (WV-2) datasets at reduced resolution.}
		\begin{tabular}{ccccc|cccc}
			\toprule
            Dataset&\multicolumn{4}{c}{QB}&\multicolumn{4}{c}{WV-2}\\
            \midrule
	        Indices&$Q_{4}$&SAM&ERGAS&SCC&$Q_{8}$&SAM&ERGAS&SCC\\
         \midrule
            Ideal value&1&0&0&1&1&0&0&1\\
                \midrule
                BT-H&0.888$\pm$0.082&2.726$\pm$0.570&6.424$\pm$1.324&0.921$\pm$0.133&0.845$\pm$0.027&6.101$\pm$1.225&4.697$\pm$0.784&0.890$\pm$0.011\\
                C-GSA&0.889$\pm$0.051&2.908$\pm$0.716&2.479$\pm$0.476&0.937$\pm$0.018&0.836$\pm$0.027&6.647$\pm$1.452&4.750$\pm$0.791&0.865$\pm$0.019\\
                SR-D&0.916$\pm$0.020&2.961$\pm$0.816&2.412$\pm$0.643&0.938$\pm$0.015&0.847$\pm$0.022&6.508$\pm$1.352&4.717$\pm$0.835&0.876$\pm$0.010\\
                MTF-GLP-FS&0.839$\pm$0.031&6.623$\pm$1.315&4.709$\pm$0.724&0.865$\pm$0.023
                &0.839$\pm$0.031&6.623$\pm$1.315&4.709$\pm$0.724&0.865$\pm$0.023\\
                PANNET&\underline{0.962$\pm$0.014}&\underline{1.770$\pm$0.436}&\underline{1.297$\pm$0.322}&\underline{0.978$\pm$0.006}&\underline{0.896$\pm$0.019}&4.868$\pm$0.948&\underline{3.943$\pm$0.695}&0.922$\pm$0.009\\
                BDPN&0.919$\pm$0.036&2.710$\pm$0.601&1.815$\pm$0.343&0.958$\pm$0.011&0.877$\pm$0.024&5.611$\pm$1.027&4.258$\pm$0.695&0.907$\pm$0.010\\
                FGFGAN&0.939$\pm$0.016&2.158$\pm$0.542&1.795$\pm$0.467&0.967$\pm$0.008&0.876$\pm$0.020&5.350$\pm$1.042&4.291$\pm$0.733&0.911$\pm$0.009\\
                NLRNET&0.910$\pm$0.034&3.004$\pm$0.558&2.815$\pm$0.279&0.955$\pm$0.012&0.889$\pm$0.027&5.886$\pm$0.772&\textbf{3.866$\pm$0.456}&\underline{0.923$\pm$0.009}\\
                TANI&0.960$\pm$0.016&1.822$\pm$0.423&1.318$\pm$0.302&\underline{0.978$\pm$0.006}&0.891$\pm$0.021&4.943$\pm$0.949&4.081$\pm$0.693&0.918$\pm$0.009\\
                \midrule
                CrossDiff$^*$&0.952$\pm$0.022&1.981$\pm$0.474&1.483$\pm$0.366&0.974$\pm$0.008&0.853$\pm$0.128&\underline{4.757$\pm$1.131}&4.044$\pm$0.705&0.921$\pm$0.008\\

                CrossDiff&\textbf{0.966$\pm$0.013}&\textbf{1.710$\pm$0.411}&\textbf{1.267$\pm$0.315}&\textbf{0.980$\pm$0.005}&\textbf{0.899$\pm$0.018}&\textbf{4.691$\pm$0.933}&3.958$\pm$0.714&\textbf{0.924$\pm$0.009}
                \\

        \bottomrule
	\end{tabular}
 \label{tb:qb_reduced}
 \small
 $^a$ CrossDiff$^*$ denotes the cross-sensor pre-trained model, \ie, pre-trained on other satellite datasets.
 \end{threeparttable}
\end{table*}


\begin{figure*}[htbp]
  \begin{tabular}{{c}{c}{c}{c}{c}{c}}
  \includegraphics[width=0.16\textwidth]{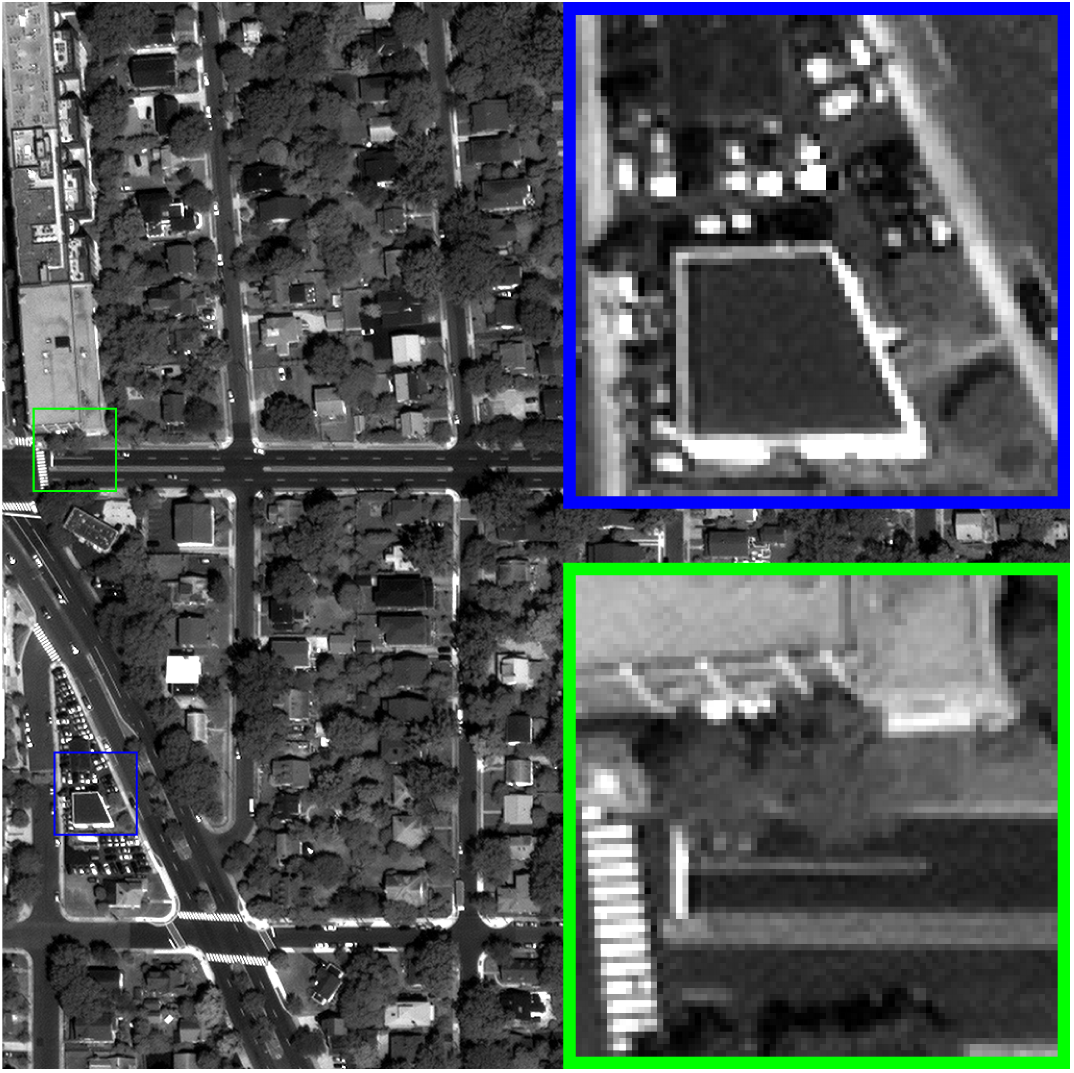} &
  \hspace{-4mm}
  \includegraphics[width=0.16\textwidth]{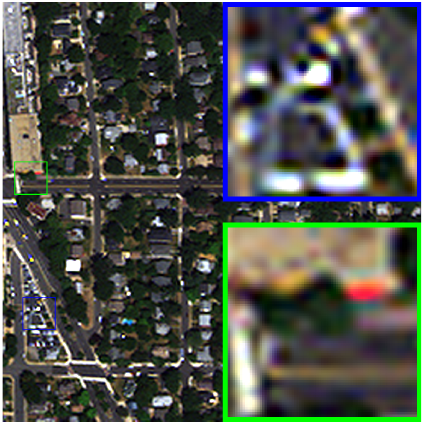} &
  \hspace{-4mm}
  \includegraphics[width=0.16\textwidth]{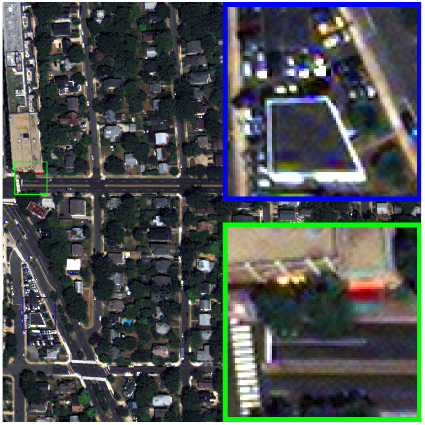} &
  \hspace{-4mm}
   \includegraphics[width=0.16\textwidth]{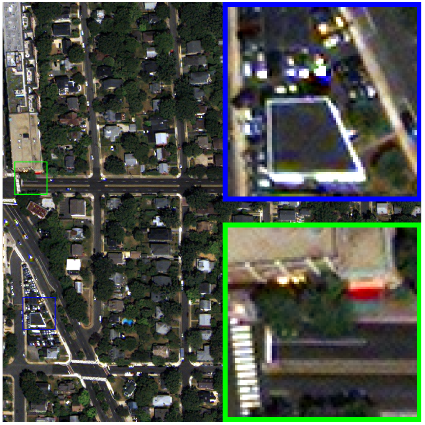}&
  \hspace{-4mm}
  \includegraphics[width=0.16\textwidth]{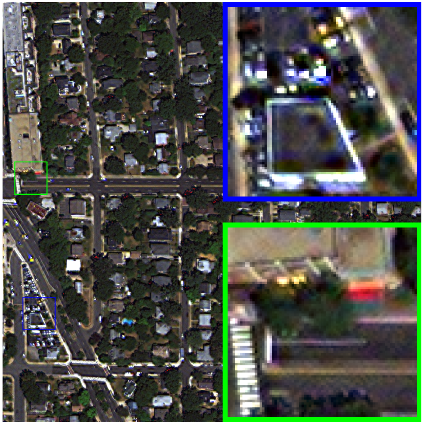}&
  \hspace{-4mm}
  \includegraphics[width=0.16\textwidth]{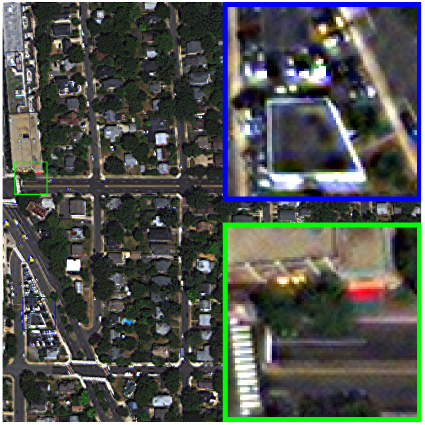}
\\

  PAN &  \hspace{-4mm}EXP &   \hspace{-4mm}BTH & \hspace{-4mm}C-GSA  &\hspace{-4mm}MTF-GLP-HPM-R&\hspace{-4mm}MTF-GLP-FS\\
  \includegraphics[width=0.16\textwidth]{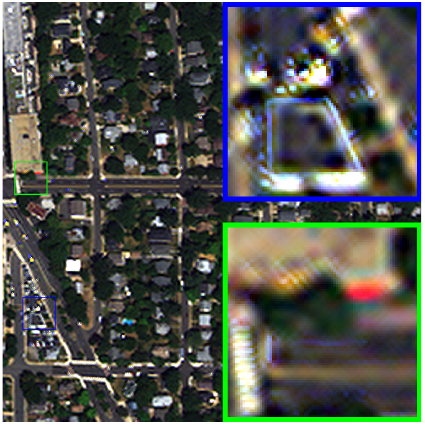}&
  \hspace{-4mm}
   \includegraphics[width=0.16\textwidth]{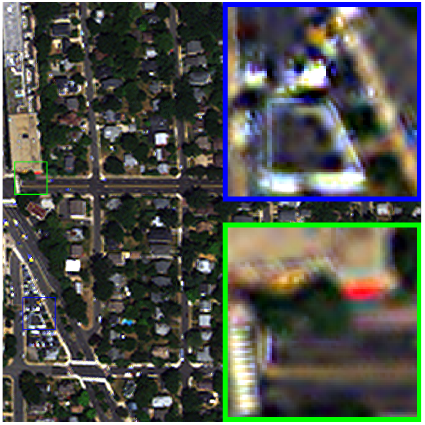}&
  \hspace{-4mm}
  \includegraphics[width=0.16\textwidth]{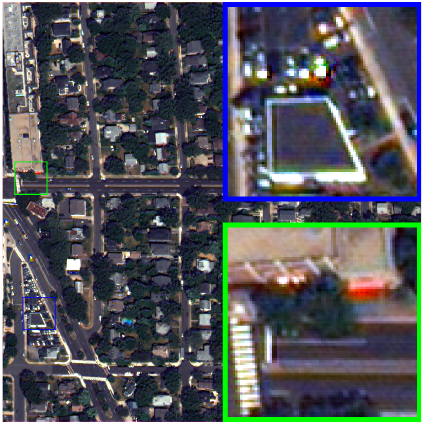}  &
  \hspace{-4mm}
  \includegraphics[width=0.16\textwidth]{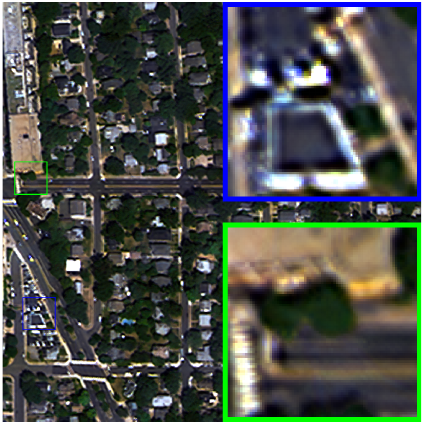}&
  \hspace{-4mm}
  \includegraphics[width=0.16\textwidth]{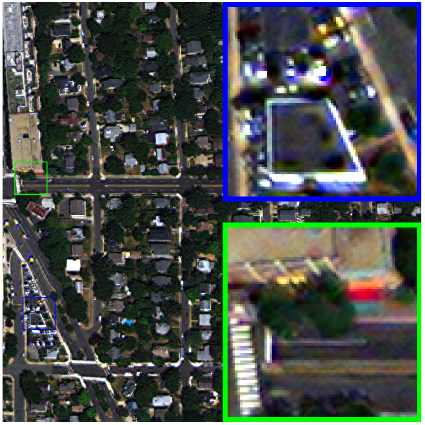}&
    \hspace{-4mm}
  \includegraphics[width=0.16\textwidth]{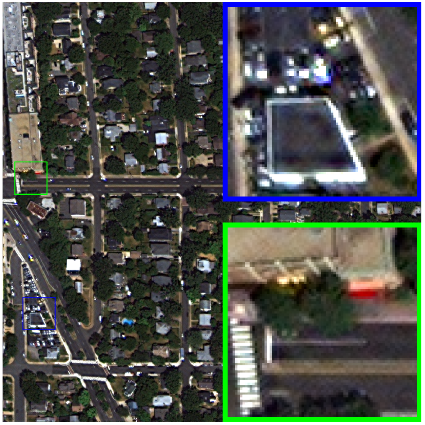}
  \\
\hspace{-4mm} ZPNN &\hspace{-4mm}  UCGAN&\hspace{-4mm} LDPNET& \hspace{-4mm}GDD& \hspace{-4mm}CrossDiff$^*$& \hspace{-4mm}CrossDiff \\

  \end{tabular} \normalsize
  \caption{Fusion results on WorldView-2 (WV-2) dataset at the full resolution. One can zoom in for more details. }
  \label{full_wv2}
\end{figure*}

\section{Experiments}
\subsection{Datasets and Evaluation Metrics}
We conduct experiments on three datasets acquired from QuickBird (QB), WorldView-4 (WV-4), and WorldView-2 (WV-2) satellites. For the page limitations, the experimental results of WV-4 dataset and the details of three datasets are provided in the supplementary material.

In the full-resolution experiments, the spectral distortion index ($D_{\lambda}$), the spatial distortion index ($D_{s}$) and the hybrid quality with no reference (HQNR)~\cite{HQNR} index are used to evaluate the quality of results, while in the reduced-resolution experiments, the widely used evaluation indexes, including SAM~\cite{yuhas1992sam}, ERGAS~\cite{wald2002ergas}, the universal image quality index for 4-band image ($Q_{4}$) and 8-band images ($Q_{8}$)~\cite{garzelli2009q2n}, and SCC~\cite{vivone2014SCC} are adopted.

\subsection{Training Details}
CrossDiff consists of a self-supervised pre-training stage and an unsupervised pansharpening adaptation stage. In both stages, we take AdamW as the optimizer with the batch size of 32 and learning rate of $3\times10^{-4}$. 
The self-supervised pre-training is conducted totally 1000 epochs. In the unsupervised adaptation stage, we first extract multi-scale features from the frozen encoders of P2M and M2P at time step 50, and then train the fusion head 20 epochs. 
Our implementation is conducted on the Nvidia GTX 3090 GPU.

\subsection{Comparisons with State-of-the-Art}

In this section, we compare our model with several state-of-the-art methods, including BT-H~\cite{lolli2017BT-H}, C-GSA~\cite{restaino2016C-GSA},  MTF-GLP-HPM-R~\cite{vivone2017MTF-GLP-HPM-R}, MTF-GLP-FS~\cite{vivone2018MTF-GLP-FS}. Because our model is an unsupervised method, we introduce extra four unsupervised DL-based methods for comparison, including ZPNN~\cite{ciotola2022ZPNN}, UCGAN~\cite{zhou2022UCGAN} and LDPNET~\cite{LDP-NET}, GDD~\cite{GDD}. Furthermore, to completely explore the self-supervised representation ability of CrossDiff, we also conduct experiments at reduced resolution, where we train the fusion head in a supervised way~\cite{wald1997wald}. Therefore, five DL-based supervised methods, \ie, PANNET~\cite{yang2017pannet}, BDPN~\cite{zhang2019BDPN}, FGFGAN~\cite{zhao2021fgfgan}, NLRNET~\cite{lei2021nlrnet} and TANI~\cite{TANI2022Diao}, are used for comparison.

\textbf{Cross-Sensor Generalization Ability}.
To verify the effectiveness of self-supervised cross-predictive pre-training, we conduct experiments to explore the model's cross-sensor generalization ability. 
Specifically, the cross-predictive diffusion model pre-trained on the WV-4 dataset is used to extract the feature representations of QB dataset, and the WorldView-3 (WV-3) pre-trained encoders are utilized to extract representations of WV-2 dataset. The results are reported in Tables~\ref{tb:full}-\ref{tb:qb_reduced}, where CrossDiff$^*$ denotes the cross-sensor pre-trained model. Note that the ZPNN, UCGAN, LDPNET and GDD in these tables are trained and tested on the same dataset. We can observe that our model can also obtain satisfactory results when the training and testing samples come from different satellite sensors, which we think due to the fact that the learned latents contain more general spatial details and spectral information after the cross-predictive diffusion pretext task. Therefore, they can be easily adapted to new satellite datasets without re-training,
which greatly boosts the cross-domain generalization ability of CrossDiff.

\textbf{Full Resolution.}
Using unsupervised loss function, we train the fusion head on original PAN and MS images.
Figures~\ref{full_qb}-\ref{full_wv2} show the visual comparisons of different methods on QB and WV-2 datasets. It can be observed that the results of UCGAN and GDD suffer from blurry effect. LDPNET and ZPNN cannot guarantee high spectral fidelity, especially for buildings. Besides, in Figure~\ref{full_qb}, BTH fails to restore spectral of the wagon, which is inconsistent with original MS image. In Figure~\ref{full_wv2}, only MTF-based traditional methods and CrossDiff have high spectral fidelity on the roof, while CrossDiff also takes spatial consistency into consideration, thus obtaining superior result. The quantitative evaluations and inference time are shown in Table~\ref{tb:full}. Both the qualitative analysis and quantitative comparisons show that CrossDiff outperforms other methods on QB and WV-2 datasets.

\textbf{Reduced Resolution.} In this experiment, we prepare the training samples according to Wald's protocol to train the fusion head by the commonly used $L1$ loss. For the page limitations, the visualizations are provided in the supplementary material. Here, we present the quantitative results in Table~\ref{tb:qb_reduced}. We can observe from Table~\ref{tb:qb_reduced} that our method outperforms the comparison methods in all metrics except for the ERGAS on WV-2 dataset. The results suggest that using the frozen encoders pre-trained by self-supervised cross-predictive diffusion process can also boost the pansharpening performance at reduced resolution. We only need to train the fusion head with a supervised loss function, which improves the flexibility of our model.

\begin{table}
\centering\caption{\protect\centering{Ablation studies on QB dataset}}
\resizebox{\columnwidth}{!}{
\begin{tabular}{c|c|c|ccc}
\toprule
  \multicolumn{2}{c}{Cross-prediction}\vline &Attention  & $D_{\lambda}$ & $D_{s}$ & HQNR \\
  \midrule
   Noise   & Clean  &  &   &   &     \\
     & &\checkmark&0.025$\pm$0.018&0.016$\pm$0.009&0.909$\pm$0.042    \\
    & \checkmark & \checkmark&\textbf{0.020$\pm$0.013}&0.027$\pm$0.012&0.892$\pm$0.047\\
 \checkmark & &  &0.030$\pm$0.015&\textbf{0.014$\pm$0.006}&0.872$\pm$0.098     \\
 \checkmark &  & \checkmark &\textbf{0.020$\pm$0.016}&0.015$\pm$0.005&\textbf{0.946$\pm$0.015}\\
\bottomrule
\end{tabular}}
\label{tb:settings of qb_full}
\end{table}

\section{Ablation Studies}\label{sc:ablation}
\textbf{Effect of Cross-Predictive Pretext Task.} In order to verify the impact of the cross-predictive pretext task on final fusion results, we random-initialize the encoders of UNet and train them together with the fusion head using the unsupervised loss.
The comparisons between the first and the last rows of Table~\ref{tb:settings of qb_full} demonstrate that the model benefits more from the cross-predictive pretext pre-training.

\textbf{Investigation on the Training Objective of Cross-Predictive Diffusion Model.}
The optimization objective of a diffusion model can either be minimization of the predicted noise~\cite{ho2020DDPM} or minimization of the reconstructed clean image~\cite{ramesh2022dalle2}. We train the cross-predictive diffusion process by using two types of training objectives. As can be seen in the second and the last rows of Table~\ref{tb:settings of qb_full}, the model that predicts noise performs better in the $D_s$ and HQNR indices, and comparable in the $D_\lambda$ index, which means that noise predictor is more suitable for our cross-predictive prediction task.

\textbf{Effect of Attention-Guided Fusion Head.}
To investigate the effect of attention guidance, we conduct experiment by removing the scSE layer in our fusion head. Experimental results are shown in the third row of Table~\ref{tb:settings of qb_full}. From the table, we can observe that attention guidance benefits the spectral preservation. Although the attention guided fusion head degrades the spatial distortion index, it improves comprehensive HQNR index by a large margin, which verifies that the attention mechanism is important for the guidance of fusion.

\section{Conclusions}\label{sc:conclution}
In this paper, we propose a cross-predictive diffusion model, dubbed CrossDiff, to explore the self-supervised representation of pansharpening. Following the forward diffusion process and the reverse denoising process of DDPM, we design a cross-predictive pretext task, where PAN image is taken as input to predict MS, and MS is inversely used to reconstruct PAN. Through such a cross-predictive self-supervised training, the encoders of the noise predictors can directly act as spectral and spatial feature extractors. By freezing the extractors, we train the fusion head on original MS and PAN image pairs in an unsupervised manner to avoid the scale variation problem.
Extensive experiments on different satellite datasets demonstrate the effectiveness of proposed method. Besides, the pre-trained diffusion model has better cross-sensor generalization ability, which can be served as effective spatial and spectral feature extractors to directly used to other satellite datasets.

\bibliographystyle{IEEEtran}
\bibliography{aaai24}

\end{document}